\documentclass{article}

\PassOptionsToPackage{numbers, compress}{natbib}
\usepackage[final]{neurips_2025}

\usepackage[utf8]{inputenc} %
\usepackage[T1]{fontenc}    %
\usepackage{hyperref}       %
\usepackage{url}            %
\usepackage{booktabs}       %
\usepackage{amsfonts}       %
\usepackage{algorithm}
\usepackage{algorithmic}
\usepackage{nicefrac}       %
\usepackage{microtype}      %
\usepackage{xcolor}  %
\usepackage{enumitem}
\usepackage{graphicx}
\usepackage{subcaption}
\usepackage{amsmath}
\usepackage{cleveref}
\usepackage{amssymb}
\usepackage{tabularx}
\usepackage{booktabs} %
\usepackage{adjustbox}
\usepackage{multirow}

\usepackage{hyperref}
\usepackage{url}
\usepackage{wrapfig}
\usepackage{enumitem}
\usepackage{multirow}
\usepackage{array}
\usepackage{makecell}
\usepackage{longtable} 
\usepackage[figuresleft]{rotating}
\usepackage{listings}
\usepackage{capt-of}
\usepackage{paracol}
\usepackage{float}

\usepackage{booktabs}
\usepackage{hyperref}
\usepackage{url}
\usepackage{wrapfig}
\usepackage{enumitem}
\usepackage{multirow}
\usepackage{array}
\usepackage{makecell}
\usepackage{longtable} 
\usepackage[figuresleft]{rotating}
\usepackage{amssymb} %
\usepackage{pifont} %

\title{Let LRMs Break Free from Overthinking \\ via Self-Braking Tuning}

\author{%
  \textbf{Haoran Zhao}\textsuperscript{1,2,*}\And 
  \textbf{Yuchen Yan}\textsuperscript{1,*}\And 
  \textbf{Yongliang Shen}\textsuperscript{1,$\dagger$}\And  
  \textbf{Haolei Xu}\textsuperscript{1}\And  
  \textbf{Wenqi Zhang}\textsuperscript{1}\AND
  \textbf{Kaitao Song}\textsuperscript{3}\And 
  \textbf{Jian Shao}\textsuperscript{1}\And 
  \textbf{Weiming Lu}\textsuperscript{1}\And 
  \textbf{Jun Xiao}\textsuperscript{1}\And 
  \textbf{Yueting Zhuang}\textsuperscript{1}\AND
  \\
  \textsuperscript{1} Zhejiang University,
  \textsuperscript{2} Tianjin University,
  \textsuperscript{3} Microsoft Research Asia \\
  \texttt{ran159753@tju.edu.cn, \{yanyuchen, syl\}@zju.edu.cn} \\
  \\
  GitHub: \url{https://github.com/ZJU-REAL/Self-Braking-Tuning} \\
  Project: \url{https://zju-real.github.io/SBT/}
}

\newcolumntype{C}[1]{>{\centering\arraybackslash}p{#1}}
\newcolumntype{L}[1]{>{\arraybackslash}p{#1}}

\begin{document}

\maketitle

\renewcommand{\thefootnote}{\fnsymbol{footnote}}
\footnotetext[1]{~The first two authors have equal contributions. This work was done when the first author was an intern at Zhejiang University.}
\footnotetext[2]{~Corresponding author.}
\renewcommand{\thefootnote}{\arabic{footnote}}

\begin{abstract}

Large reasoning models (LRMs), such as OpenAI o1 and DeepSeek-R1, have significantly enhanced their reasoning capabilities by generating longer chains of thought, demonstrating outstanding performance across a variety of tasks. However, this performance gain comes at the cost of a substantial increase in redundant reasoning during the generation process, leading to high computational overhead and exacerbating the issue of overthinking.
Although numerous existing approaches aim to address the problem of overthinking, they often rely on external interventions.
In this paper, we propose a novel framework, \textbf{Self-Braking Tuning} (SBT), which tackles overthinking from the perspective of allowing the model to regulate its own reasoning process, thus eliminating the reliance on external control mechanisms. We construct a set of overthinking identification metrics based on standard answers and design a systematic method to detect redundant reasoning. This method accurately identifies unnecessary steps within the reasoning trajectory and generates training signals for learning self-regulation behaviors. Building on this foundation, we develop a complete strategy for constructing data with adaptive reasoning lengths and introduce an innovative braking prompt mechanism that enables the model to naturally learn when to terminate reasoning at an appropriate point.
Experiments across mathematical benchmarks (AIME, AMC, MATH500, GSM8K) demonstrate that our method reduces token consumption by up to 60\% while maintaining comparable accuracy to unconstrained models. 
\end{abstract}

\begin{figure}[h]
  \centering
  \includegraphics[width=\textwidth]{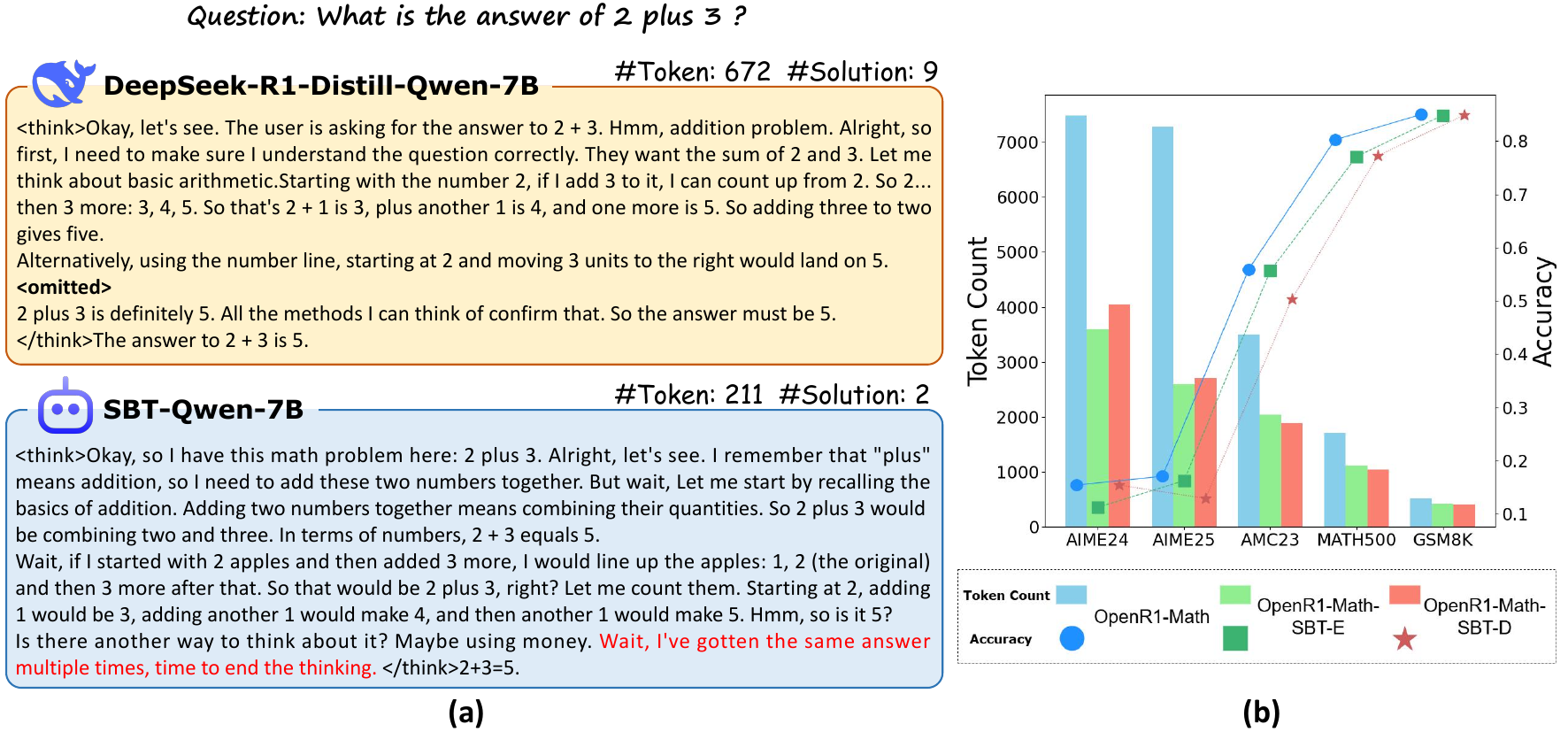}
  \caption{Demonstration of Self-Braking Tuning Effectiveness. In the single-example case (a), the self-braking tuned model exhibits spontaneous termination of overthinking and significantly reduces token usage. On major mathematical benchmarks (b), compared to using OpenR1-Math~\cite{openr1math220k} as the SFT dataset, the self-braking tuned Qwen2.5-Math-1.5B-Instruct~\cite{yang2024qwen2} achieves a substantial reduction in tokens consumed during inference while maintaining comparable accuracy.}
  \label{fig:1}
  \vspace{-1.5em}
\end{figure}

\section{Introduction}

Large reasoning models (LRMs) such as OpenAI’s o1~\cite{jaech2024openai}, Deepseek-R1~\cite{guo2025deepseek}, QwQ~\cite{qwen2024qwq}, Gemini 2.0 Flash Thinking~\cite{gemini_flash_thinking} and Kimi-1.5~\cite{team2025kimi}, excel at mathematical and logical tasks by generating detailed multi-step reasoning, boosting accuracy on complex benchmarks~\cite{latif2024can}. However, this often results in excessively long inference trajectories, frequently consuming thousands of tokens per problem~\cite{learning_to_reason_with_llms,wang2025thoughts}, leading to increased computational cost, latency, and redundant reasoning that can obscure core solutions~\cite{wumore}. This “overthinking”~\cite{chen2024not} poses a significant challenge for practical deployment.

Many recent studies have focused on addressing the problem of overthinking~\cite{sui2025stop,wang2025harnessing}, which can be broadly categorized into three approaches: (1) \textit{Model optimization:} Apply reinforcement learning (RL) or supervised fine-tuning (SFT) to equip models with the ability to control reasoning length~\citep{aggarwal2025l1controllinglongreasoning,luo2025o1,yeo2025demystifying}; (2) \textit{Reasoning output optimization}: Dynamically reducing the number of reasoning steps and output length during inference~\cite{ma2025reasoning,yang2025dynamic,zhang2025lightthinker}; (3) \textit{Adding external restrictions}: Imposing external constraints, such as token budgets, to reduce overthinking~\cite{han2024token,xu2025chain}. Most existing methods follow the paradigm of external intervention, relying on complex optimization strategies or introducing additional constraint mechanisms, and have yet to fully explore the intrinsic ability of the model to mitigate overthinking on its own.

This reliance on external control prompts a fundamental question: \textbf{\textit{Can we enable large reasoning models to autonomously recognize excessive reasoning and terminate their thinking process appropriately?}} Ideally, a model should intrinsically understand when additional reasoning becomes redundant and halt its thought process without external triggers, similar to how humans naturally conclude their reasoning when reaching sufficient certainty.

To address this challenge, we propose \textbf{Self-Braking Tuning (SBT)}, a novel framework that teaches LRMs to autonomously identify and terminate redundant reasoning. Unlike previous approaches that impose external constraints, SBT fundamentally reshapes how models perceive and regulate their own reasoning processes. Our key insight is that LRMs can be trained to develop an internal braking mechanism that recognizes when further reasoning becomes unproductive, enabling them to naturally conclude the thought process and transition to formulating the final solution.

Our approach begins with a systematic methodology for identifying overthinking patterns in reasoning trajectories. By combining metrics such as reasoning efficiency ratio and overthinking label ratio, we precisely pinpoint the transition point at which the model shifts from effective reasoning to redundant computation. Based on this analysis, we develop two complementary data construction strategies: (1) Self-Braking Tuning Exact (SBT-E), which strictly removes redundant reasoning segments based on predefined braking points, allowing the model to enter the conclusion phase earlier; (2) Self-Braking Tuning Dynamic (SBT-D), which implements step-level monitoring and dynamically halts reasoning when overthinking patterns emerge.

Building upon the high-quality OpenR1-Math~\cite{openr1math220k} dataset of reasoning trajectories, we constructed two specialized training datasets using these strategies: OpenR1-Math-SBT-E and OpenR1-Math-SBT-D. To further enhance the model's self-awareness of its reasoning state, we introduce braking prompts at the identified braking points, explicitly simulating the recognition of having sufficiently completed the reasoning process. These prompts enable the model to naturally express an awareness of having reached adequate reasoning depth, thus promoting autonomous termination without the need for external signals. Our experimental results demonstrate consistent improvements in reasoning efficiency across mathematical benchmarks of varying difficulty. While maintaining high accuracy, token consumption was reduced by 30\% to 60\%. 

Our contributions can be summarized as follows:

\begin{itemize}[leftmargin=6mm]
    \item We introduce a novel tuning framework that enables LRMs to self-regulate reasoning length without external constraints. Self-Braking Tuning cultivates models' intrinsic ability to recognize and inhibit excessive reasoning, fundamentally improving inference efficiency and response quality.
    \item We propose a systematic methodology for identifying overthinking patterns and develop two complementary data construction strategies: SBT-E and SBT-D, resulting in specialized training datasets for addressing the overthinking problem. These datasets systematically prune redundant reasoning while preserving essential thinking steps.
    \item We demonstrate that models trained with our SBT framework maintain original accuracy levels while reducing token consumption by up to 60\% across multiple benchmarks, confirming the effectiveness and generalizability of our approach in enhancing reasoning efficiency.
\end{itemize}

\section{Related Works}
\paragraph{Large reasoning models}
Large reasoning models (LRMs) extend traditional language models with advanced reasoning capabilities, often enabled by reinforcement learning. OpenAI’s o1 series~\cite{jaech2024openai} marked a key milestone, followed by DeepSeek-R1~\cite{guo2025deepseek}, which matched o1's performance through a cost-efficient combination of supervised fine-tuning and RL. Subsequent models like Kimi-k1.5~\cite{team2025kimi} and QwQ-32B~\cite{qwen2024qwq} further solidified the LRM era. Concurrently, alternative approaches have emerged to reduce reliance on RL, leveraging supervised fine-tuning and data distillation. Models such as DeepSeek-Distill~\cite{guo2025deepseek}, OpenR1-Math-7B~\cite{openr1}, Sky-T1~\cite{team2025sky}, and LIMO~\cite{ye2025limo} have shown that strong reasoning performance can also be achieved without RL, broadening the design space for LRMs. In addition, recent algorithmic advances specifically target robustness and scalability of reasoning~\cite{xu2025mind,yan2025mathfimer,xu2025easysteer}: S$^3$c-Math~\cite{yan2025s} introduces spontaneous step-level self-correction to let models detect and fix erroneous intermediate steps during chain-of-thought generation, while InftyThink~\cite{yan2025inftythink} breaks long-context limits by turning monolithic proofs into iterative short-segment reasoning with concise progress summaries, enabling effectively unbounded reasoning depth with bounded computation.

\paragraph{Efficient reasoning}

Overthinking is a common behavior of large reasoning models, where models generate unnecessarily long answers instead of stopping at the right time. Existing studies addressing overthinking~\cite{sui2025stop} can be grouped into three categories: (1) \textit{Model Optimization:} This line improves model behavior via post-training techniques, primarily reinforcement learning (RL) and supervised fine-tuning (SFT). RL-based methods design length-sensitive rewards to limit output~\cite{luo2025o1,yeo2025demystifying,aggarwal2025l1controllinglongreasoning,shen2025dast}. SFT-based methods use variable-length chain-of-thought (CoT) data and auxiliary constraints to shorten reasoning paths~\cite{yu2distilling,munkhbat2025self,kang2025c3ot,xia2025tokenskip,ma-etal-2025-cot}. (2) \textit{Reasoning Output Optimization:} This direction reduces reasoning length at inference time by altering generation strategies. \textit{lightthinker}~\cite{zhang2025lightthinker} compresses intermediate steps; \textit{DEER}~\cite{yang2025dynamic} halts once high confidence is reached; \textit{NoThinking}~\cite{ma2025reasoning} skips reasoning entirely via prompting. These efforts are representative, with many other studies exploring similar directions~\cite{liaoreward,liescape,manvi2024adaptive,zhang2025reasoning}. (3) These methods impose constraints like token budgets or prompt controls to regulate reasoning behavior. In addition to representative works such as Token-Budget~\cite{han2024token} and CoD~\cite{xu2025chain}, numerous other studies have explored similar constraint-based strategies~\cite{lee2025well,renze2024benefits,aytes2025sketch}.

\section{Methods}
\label{methods}

\begin{figure}[t]
\centering
\includegraphics[width=\textwidth]{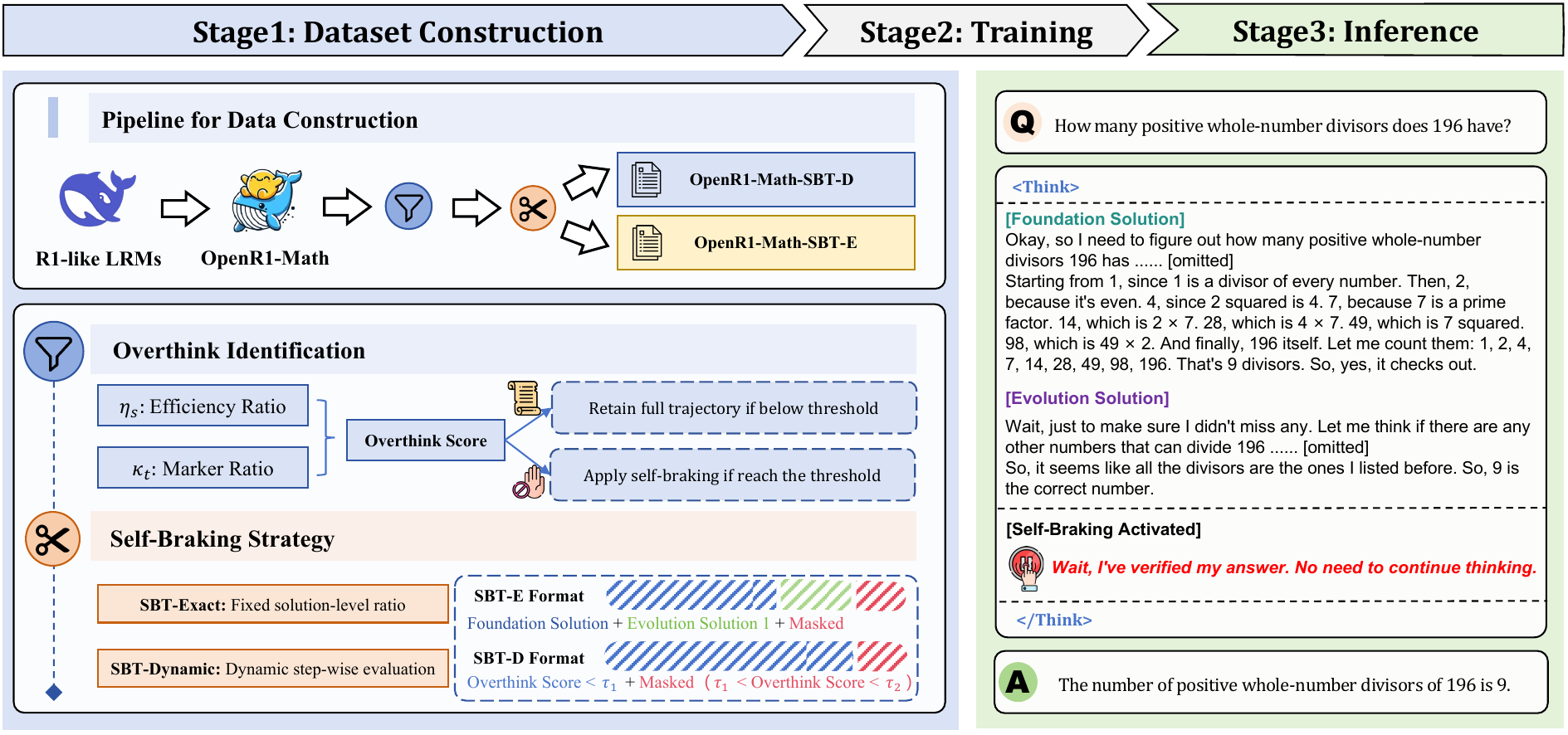}
\caption{Overview of Self-Braking Tuning. Left: Data construction process with overthinking identification and self braking truncation strategies. Right: An example of automatic reasoning termination in a trained Self-Braking LLM.}
\label{fig:overview}
\end{figure}

In this section, we begin by analyzing the reasoning trajectories of LRMs to understand the patterns of overthinking (Section \ref{sec:Trajectory}). Based on this analysis, we propose metrics to quantify overthinking (Section \ref{sec:overthinking}). We then introduce our Self-Braking Tuning framework, which includes two data construction strategies (Section \ref{sec:data}) and a braking prompt mechanism (Section \ref{sec:warning}).

\subsection{Reasoning trajectory analysis in R1-like models}
\label{sec:Trajectory}

\begin{figure}[t]
  \centering
  \begin{subfigure}[b]{0.59\textwidth}
    \includegraphics[width=\textwidth]{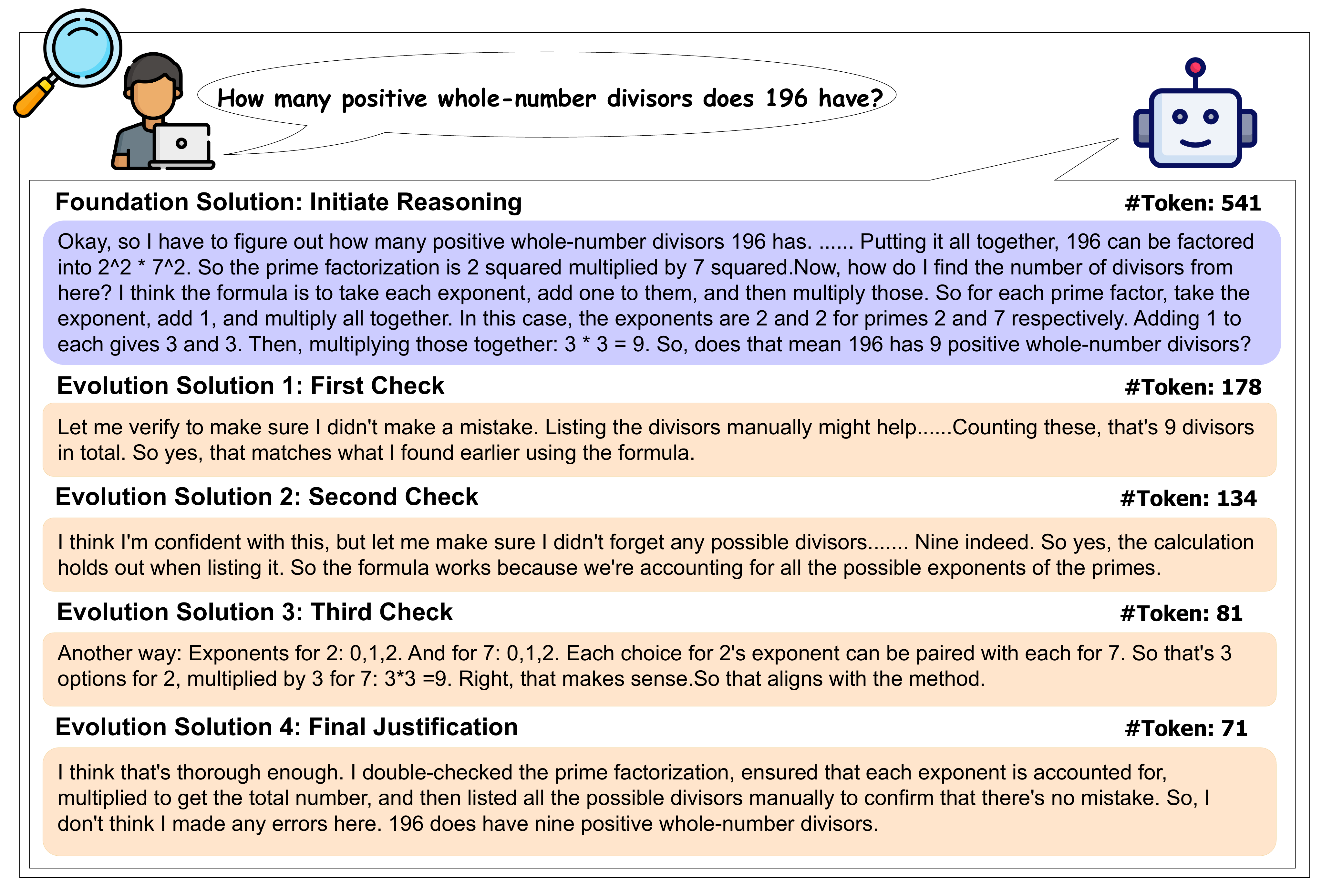}
    \caption{}
    \label{fig:thought}
  \end{subfigure}
  \begin{subfigure}[b]{0.39\textwidth}
    \includegraphics[width=\textwidth]{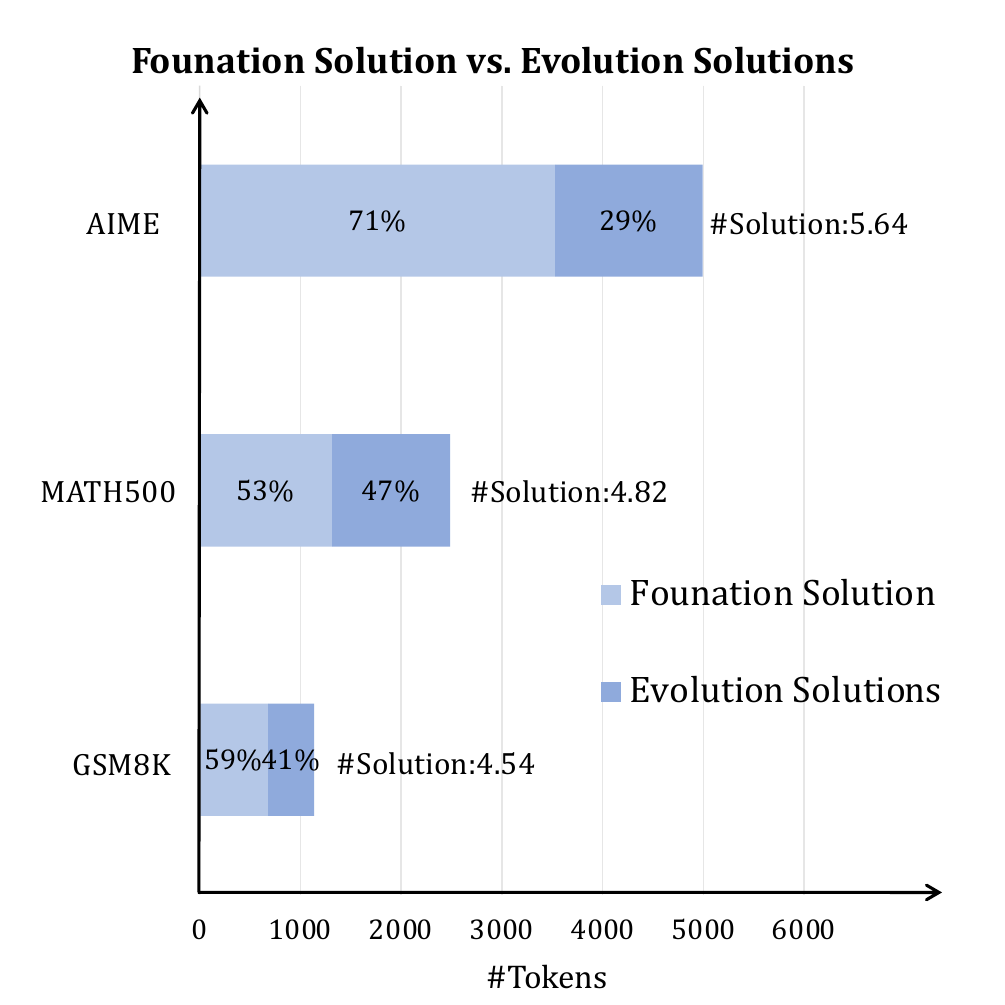}
    \caption{}
    \label{fig:fvse}
  \end{subfigure}
  \caption{In panel (a), we present a representative example from DeepSeek-R1-Distill-Qwen-7B. In panel (b), we analyze the model’s performance across GSM8K, MATH500, and AIME benchmarks, showing that the Foundation Solution plays a critical role across tasks of varying difficulty.}
  \label{fig:2}
  \vspace{-12pt}
\end{figure}

Understanding the reasoning structure of LRMs is key to addressing overthinking. Analysis of trajectories from models like DeepSeek-R1 reveals a common pattern: multiple distinct solution attempts are often generated for a single problem. Based on their role and position, these solution segments can be grouped into two main types:

\begin{enumerate}[leftmargin=6mm, label=\arabic*.]
    \item \textbf{Foundation solution:} This is the first solution at the beginning of its reasoning process. After comprehending the problem, it proceeds with a step-by-step solution. This forms the foundation of the reasoning process and guides the development of subsequent Evolution Solutions.
    \item \textbf{Evolution solution:} These solutions appear in the later stages of the model’s reasoning process and are often introduced with cues such as “Wait,” “Alternatively,” or “However.” Evolution solutions primarily reflect, refine, supplement, or summarize the foundational solution, and may propose new approaches. While this part of the reasoning grants the model self-correction and improvement capabilities, it is also where overthinking most frequently occurs.
\end{enumerate}
To illustrate how LRMs behave across varying difficulty levels, we constructed a set of math evaluation benchmarks with graduated difficulty and conducted experiments on DeepSeek-Distill-Qwen-7B. The distribution of correct reasoning trajectories is reported in Figure \ref{fig:2}. Additionally, a representative example from the MATH500~\cite{hendrycks2021measuring,lightman2023let} task is presented to demonstrate the concrete forms of Foundation Solution and Evolution Solution.

\subsection{Overthinking identification}
\label{sec:overthinking}
Identifying overthinking in reasoning trajectories is crucial for designing effective self-braking mechanisms. Based on our structural analysis of reasoning, we propose a quantitative framework for detecting and measuring redundancy in the OpenR1-Math dataset, which spans a range of mathematical problems and difficulty levels. To distinguish essential reasoning from unnecessary computation, we introduce two complementary metrics capturing different forms of redundancy, integrated through a composite scoring mechanism.
\paragraph{Reasoning efficiency ratio}
Our first metric addresses a fundamental observation in Section \ref{sec:Trajectory}: LRMs frequently derive correct answers relatively early in their reasoning process but continue generating additional solution attempts. To quantify this inefficiency, we introduce the reasoning efficiency ratio, denoted as
\begin{equation}
\label{equ:rer}
\eta_{s} = \frac{FS}{TS}
\end{equation}
where \(FS\) (first correct steps) represents the number of reasoning steps required to reach the first correct answer within the thinking segment (enclosed by \texttt{<think>} and \texttt{</think>} tags) and \(TS\) (total thinking steps) represents the total number of steps in the entire thinking segment.

This ratio provides a direct measure of reasoning efficiency: values closer to 1 indicate that the model spent most of its reasoning steps before arriving at the correct answer, reflecting a focused and efficient reasoning process.  In contrast, values closer to 0 suggest that the model continued reasoning extensively after reaching the correct answer, which may indicate overthinking or redundant computation.  The step-based calculation enables us to assess the structural efficiency of the reasoning process independently of implementation-specific token counts.

\paragraph{Overthinking marker ratio} 

While $\eta_s$ captures structural inefficiency, it does not account for linguistic patterns characteristic of overthinking. Through analysis of high-quality DeepSeek-R1 reasoning trajectories, we identified a set of linguistic markers strongly associated with overthinking behaviors---terms that signal reconsideration, verification, or alternative approach exploration. We formalize this observation through the overthinking marker ratio, denoted as $\kappa_t$:

\vspace{-1em}
\begin{equation}
\kappa_{t} = \frac{1}{TT} \sum_{i=1}^{TT} \mathbb{I}[w_i \in \mathcal{M}], \quad \mathbb{I}[w_i \in \mathcal{M}]=
\begin{cases}
    1, & \text{if } w_i \in \mathcal{M} \\
    0, & \text{otherwise}
\end{cases} \\
\end{equation}

where $\mathcal{M}$ represents our curated lexicon of overthinking marker terms (complete list in Appendix \ref{markers}), TT (Total Tokens) represents the total number of tokens in the thinking segment, $\mathbb{I}[\cdot]$ is the indicator function that equals 1 when $w_i$ belongs to $\mathcal{M}$ and 0 otherwise.

This ratio quantifies the linguistic footprint of overthinking within a reasoning trajectory. Higher values of $\kappa_t$ indicate a greater presence of reconsideration and verification language, which typically correlates with redundant reasoning patterns.
\paragraph{Overthink score} 
To develop a comprehensive assessment of overthinking that leverages both structural and linguistic indicators, we introduce the overthink score, a weighted combination of our two metrics:

\vspace{-1em}
\begin{equation}
\text{Overthink\hspace{0.33em}Score} = \beta \times \kappa_{t} + (1 - \beta) \times (1 - \eta _{s})
\end{equation}

Note that we transform $\eta_s$ to $(1 - \eta_s)$ to ensure directional consistency, as higher values of $\eta_s$ indicate greater efficiency (less overthinking), while higher values of $\kappa_t$ suggest stronger overthinking patterns. The weighting parameter $\beta \in [0,1]$ balances the contribution of each component to the final score.

In our implementation, we set $\beta = 0.1$ based on both theoretical considerations and empirical validation. This parameter choice reflects two key insights:

\begin{itemize}[leftmargin=6mm]
    \item \textbf{Reasoning efficiency dominance}: Early arrival at correct answers significantly reduces computational resources and latency, making $(1 - \eta_s)$ the primary component (weighted at 90\%). This prioritization aligns with our objective of minimizing unnecessary computation while preserving reasoning quality.
    \item \textbf{Linguistic indicator robustness}: While $\kappa_t$ provides valuable signals about overthinking, it exhibits greater sensitivity to variations in prompt formulation, corpus characteristics, and model-specific output patterns. Assigning a lower weight (10\%) to $\kappa_t$ mitigates potential noise amplification from these stylistic fluctuations.
\end{itemize}

\subsection{Adaptive inference data construction}
\label{sec:data}

Building on our quantitative framework for overthinking identification, we introduce two complementary strategies for constructing adaptive inference length datasets: Self-Braking Tuning Exact (SBT-E) and Self-Braking Tuning Dynamic (SBT-D). Both approaches aim to preserve reasoning depth while cultivating the model's ability to terminate excessive thinking.

\paragraph{Self-Braking Tuning Exact}
SBT-E implements a solution-level truncation strategy with consistent reasoning structure across all training examples. For each trajectory exhibiting overthinking, we preserve the Foundation Solution plus one Evolution Solution, followed by a small masked segment from subsequent reasoning. This structured approach ensures the model learns clear boundaries between necessary reasoning and excessive computation.

The Foundation Solution captures the initial structured approach to the problem, while the additional Evolution Solution preserves self-correction capabilities. The masked segment, comprising the beginning of the next Evolution Solution, serves as a braking indicator that signals where further reasoning becomes redundant. During training, this masked content does not contribute to the loss function, thereby avoiding reinforcement of overthinking patterns.
Algorithm \ref{alg:sbt-e} in Appendix presents the formal procedure for SBT-E construction.

\vspace{-5pt}
\paragraph{Self-Braking Tuning Dynamic}
While SBT-E uses uniform truncation, SBT-D adopts a step-wise adaptive strategy that tailors reasoning length to each problem. It incrementally analyzes each reasoning step to determine individualized termination points.

The process begins with the full preservation of the Foundation Solution. Subsequent steps are added one by one, with the overthink score recalculated after each. Reasoning continues until the score surpasses a primary threshold $\tau_1$ (set to 0.2), allowing complex problems to retain more steps and simpler ones to terminate earlier.

A masking segment is then defined from steps with overthink scores between $\tau_1$ and a secondary threshold $\tau_2$ (set to $\tau_1 + 5\%$). This segment is excluded from loss computation but retained to expose the model to overthinking patterns without reinforcing them. The full procedure is outlined in Algorithm~\ref{alg:sbt-d} in Appendix.

Together, SBT-E and SBT-D yield the OpenR1-Math-SBT-E and OpenR1-Math-SBT-D datasets, each with 92,064 examples, designed to train models to autonomously terminate redundant reasoning.

\subsection{Self-regulating braking strategy} 
\label{sec:warning}
Beyond generating adaptive-length reasoning, we introduce complementary training mechanisms to enhance the model’s ability to stop reasoning autonomously. Our Self-Regulating Braking Strategy includes two components: masked redundant thinking and natural language braking signals—both aimed at fostering self-awareness of reasoning efficiency.

\paragraph{Masked redundant thinking}

While both SBT-E and SBT-D identify optimal truncation points, simply cutting off reasoning there doesn’t help models learn to detect overthinking. Instead, we retain a small portion of redundant reasoning and apply loss masking to prevent it from affecting training. For each SBT-E or SBT-D sample, we append this masked segment right after the preserved valid reasoning: in SBT-E, it’s the start of the second Evolution Solution; in SBT-D, it includes steps with overthink scores between thresholds $\tau_1$ and $\tau_2$. This consistent strategy across both methods ensures balanced exposure.

By presenting overthinking patterns without calculating their loss, the model learns to distinguish between productive reasoning and redundancy. The masked segment serves as a soft boundary cue, encouraging the model to stop reasoning autonomously during inference.

\paragraph{Natural language guidance}
We further enhance self-regulation by adding clear natural language cues at reasoning stop points. These braking signals are self-reflective statements that show awareness of finishing reasoning. For example, \textit{“Wait, I've gotten the same answer multiple times, time to end the thinking.”}

Placed at the boundary between preserved and masked content, these cues act as linguistic anchors for stopping decisions. Unlike special tokens or external rules, natural language signals fit naturally with the model’s abilities, provide explicit metacognitive hints, and keep the reasoning fluent while clearly indicating when to stop.

\section{Experiments}
\label{ex}

We conduct extensive experiments to evaluate the effectiveness of Self-Braking Tuning across various model architectures and mathematical reasoning tasks. Our evaluation aims to answer three key questions: (1) How effectively does SBT reduce token consumption while preserving accuracy? (2) How does performance vary across different model sizes and architectures? (3) How do the two SBT variants (SBT-E and SBT-D) compare in practice?

\subsection{Experimental setup}
\label{subsec:setup}

\paragraph{Datasets and training.} We curate a dataset of 92K high-quality instances from OpenR1-Math~\cite{openr1math220k}\footnote{The OpenR1-Math dataset is licensed under the Apache 2.0 License. We adhere to its terms of use and do not redistribute the dataset but build upon it for experimental purposes.} by applying a 16K context limit and filtering out problematic samples (e.g., those with multiple \texttt{</think>} tags). This filtered dataset serves as our baseline. We then construct our SBT-E and SBT-D variants following the methodology detailed in Section~\ref{sec:data}.

We perform supervised fine-tuning on both mathematical specialists (Qwen2.5-Math-1.5B/7B-Instruct~\cite{yang2024qwen2}) and general-purpose models (Llama-3.2-1B and Llama-3.1-8B-Instruct~\cite{grattafiori2024llama}). All models are trained using Megatron-LM for 3 epochs with a 1e-5 initial learning rate, cosine decay schedule, 0.03 warm-up ratio, and 16,384-token maximum sequence length. Training is conducted on 64 Ascend H910B-64G hardware.

\paragraph{Evaluation benchmarks.} We evaluate performance across four mathematical reasoning benchmarks of varying difficulty: AIME (24\&25, competition-level algebraic problems), AMC23 (pre-collegiate mathematics), MATH500~\cite{hendrycks2021measuring, lightman2023let} (diverse mathematical problems), and GSM8K~\cite{cobbe2021training} (grade school math word problems). For inference, we use vLLM~\cite{kwon2023efficient} with temperature 0.7, generating 8 samples per question and reporting Average accuracy. All inference are performed on NVIDIA A100 GPUs.

\begin{table}[t]
\centering
\small
\caption{Performance of different models with Self-Braking Tuning applied, evaluated across GSM8K, MATH500, AMC23, and AIME (including AIME24 and AIME25) benchmarks.}
\label{tab:main}
\vspace{5pt}
\begin{adjustbox}{max width=\textwidth}
\begin{tabular}{C{1.8cm}C{1cm} 
C{0.6cm}C{0.6cm}
C{0.6cm}C{0.6cm}
C{0.6cm}C{0.6cm}
C{0.6cm}C{0.6cm}
C{0.6cm}C{0.6cm}}
\toprule
\multirow{2}{*}{\centering \textbf{Base Model}} & \multirow{2}{*}{\centering \textbf{Method}}
& \multicolumn{2}{c}{\textbf{GSM8K}} 
& \multicolumn{2}{c}{\textbf{MATH500}} 
& \multicolumn{2}{c}{\textbf{AIME}} 
& \multicolumn{2}{c}{\textbf{AMC23}} 
& \multicolumn{2}{c}{\textbf{Average}}\\
\cmidrule(lr){3-4} \cmidrule(lr){5-6} \cmidrule(lr){7-8} \cmidrule(lr){9-10} \cmidrule(lr){11-12}
& & Acc & \#Tok & Acc & \#Tok & Acc & \#Tok & Acc & \#Tok & Acc & \#Tok \\
\midrule
\multirow{3}{*}{\shortstack{Qwen2.5-Math-\\1.5B-Instruct}} & Baseline & 85.00 & 514 & 80.25 & 1712 & 16.25 & 7381 & 55.94 & 3503 & \textbf{59.36} & 3277 \\
& SBT-E       & 84.85 & 426 & 77.10 & 1121 & 13.75 & 3101 & 55.63 & 2044 & 57.83 & \textbf{1673} \\
& SBT-D       & 84.87 & 414 & 77.30 & 1046 & 14.17 & 3381 & 50.31 & 1888 & 56.66 & 1682 \\
\midrule
\multirow{3}{*}{\shortstack{Qwen2.5-Math-\\7B-Instruct}} & Baseline & 96.11 & 1460 & 92.67 & 3816 & 40.83 & 11904 & 83.13 & 6937 & \textbf{78.19} & 6029 \\
& SBT-E     & 95.45 & 997 & 90.77 & 2501 & 38.75 & 8772  & 77.19 & 4443 & 75.54 & \textbf{4178}  \\
& SBT-D     & 95.37 & 956 & 91.15 & 2629 & 38.38 & 9778 & 80.06 & 5208 & 76.24 & 4643 \\
\midrule
\multirow{3}{*}{\shortstack{Llama-3.2-\\1B-Instruct}} & Baseline & 41.85 & 1639 & 25.22 & 6624 & 1.25 & 13150 & 9.38 & 10210 & 19.43 & 7906\\
& SBT-E    & 39.96 & 1056 & 24.35 & 3180 & 0.42 & 6615  & 9.06 & 4708 & 18.45 & 3890 \\
& SBT-D    & 41.21 & 698 & 25.07 & 2591 & 1.04 & 6821  & 13.13 & 4388 & \textbf{20.11} & \textbf{3624} \\
\midrule
\multirow{3}{*}{\shortstack{Llama-3.1-\\8B-Instruct}} & Baseline & 88.03 & 1593 & 59.98 & 9304 & 9.58 & 13663 & 36.75 & 9742 & 48.59 & 8576 \\
& SBT-E         & 85.03 & 777 & 57.60 & 2292 & 6.84 & 5658 & 33.44 & 4045 & 45.73 & \textbf{3193} \\
& SBT-D         & 88.27 & 997 & 62.60 & 3847 & 7.70 & 5845 & 38.12 & 6476 & \textbf{49.17} & 4291\\
\bottomrule
\end{tabular}
\end{adjustbox}
\vspace{-20pt}
\end{table}

\subsection{Main results}
\label{subsec:main_results}
Table~\ref{tab:main} presents the performance of models trained with our Self-Braking Tuning approaches compared to baselines trained on the unmodified dataset. We observe several significant trends:
\paragraph{Substantial token reduction with preserved acc.} Both SBT variants achieve remarkable reductions in token consumption while maintaining comparable accuracy to baseline models. For the Qwen2.5-Math-7B-Instruct model, SBT-E and SBT-D reduce token usage by 30.7\% and 23.0\% respectively, with accuracy drops of only 2.65\% and 1.95\%. Even more impressively, when applied to the Llama-3.1-8B-Instruct model, SBT-E reduces token consumption by 62.8\% while preserving 94.1\% of the baseline accuracy.
\paragraph{Scaling dynamics across model types} Efficiency gains from Self-Braking Tuning (SBT) vary by model type.  For general-purpose models like Llama, larger models benefit more---token reductions improve from 54.2\% (1B) to 62.8\% (8B).  In math-specialized models, however, larger models see smaller gains (30.7\% for 7B vs. 48.9\% for 1.5B), suggesting that specialized models already have more focused and efficient reasoning, leaving less room for further compression. These findings indicate that the self-regulation benefits from SBT depend not only on model scale but also on whether the model is trained for general-purpose or domain-specific tasks.
\paragraph{SBT-E vs. SBT-D performance.} The two proposed variants show distinct performance characteristics. SBT-E generally achieves greater token reductions (averaging 48.3\% across all models compared to 43.9\% for SBT-D) but with slightly larger accuracy drops. SBT-D demonstrates more balanced performance, particularly on the most challenging AIME and MATH500 benchmarks. Notably, for the Llama-3.1-8B model, SBT-D actually improves accuracy on MATH500 by 2.62 \% while reducing tokens by 58.7\%, suggesting that dynamic truncation may help eliminate not just redundant reasoning but potentially harmful overthinking in some cases.

\section{Analysis}
\label{sec:analysis}

\subsection{Impact of overthinking thresholds}
\label{subsec:thresholds}

The threshold for classifying overthinking instances significantly impacts both dataset composition and model performance. We experimented with thresholds of 0.2, 0.3, and 0.4, which classified approximately 60\%, 50\%, and 40\% of samples as overthinking cases, respectively. As shown in Table~\ref{tab:threshold}, a threshold of 0.2 yields the best performance for SBT-E, achieving an optimal balance between token reduction (49\% fewer tokens than baseline) and accuracy preservation (97.4\% of baseline accuracy).

This finding reveals a crucial insight: aggressive overthinking identification (lower thresholds) leads to more substantial efficiency gains without proportional accuracy losses. This suggests that a significant portion of reasoning in LRMs truly is redundant and can be eliminated without compromising problem-solving capabilities. The consistent pattern across both SBT variants indicates that identifying and addressing overthinking in approximately 60\% of reasoning instances represents an optimal operating point for Self-Braking Tuning.

\begin{table}[t!]
    \centering
    \begin{minipage}[t]{0.38\linewidth}
        \centering
        \small
        \caption{Performance across overthink score thresholds. Detailed results shown in Appendix Table~\ref{tab:rawof5.1}.}
        \label{tab:threshold}
        \vspace{4pt}
        \resizebox{\linewidth}{!}{
            \begin{tabular}{l c c c}
                \toprule
                \textbf{Method} & \textbf{Threshold} & \textbf{Acc} & \textbf{\#Tok} \\
                \midrule
                Baseline      & --   & 59.36 & 3278 \\
                \midrule
                SBT-Exact     & 0.2  & \textbf{57.83} & \textbf{1673} \\
                              & 0.3  & 56.70 & 1755 \\
                              & 0.4  & 57.38 & 1834 \\
                \midrule
                SBT-Dynamic   & 0.2  & 56.66 & \textbf{1682} \\
                              & 0.3  & \textbf{57.47} & 1917 \\
                              & 0.4  & 57.36 & 1902 \\
                \bottomrule
            \end{tabular}
        }
    \end{minipage}
    \hspace{2mm}
    \begin{minipage}[t]{0.57\linewidth}
        \centering
        \small
        \caption{Performance with different configurations of preserved reasoning and masked redundant content. Detailed results shown in Appendix Table~\ref{label:rawof5.2}.}
        \label{tab:masking}
        \vspace{4pt}
        \resizebox{\linewidth}{!}{
            \begin{tabular}{lcc}
                \toprule
                \textbf{Reservations \& Masked Content} & \textbf{Acc} & \textbf{\#Tok} \\
                \midrule
                Baseline & 59.36 & 3277 \\
                \midrule
                1 solution \& A few sentences & 56.95 & 1700 \\
                1 solution \& 1 solution & 57.69 & 1697 \\
                2 solutions \& A few sentences & \textbf{57.83} & \textbf{1673} \\
                2 solutions \& 1 solution & 57.45 & 1684 \\
                \bottomrule
            \end{tabular}
        }
    \end{minipage}
    \vspace{-16pt}
\end{table}

\subsection{Preserved reasoning and redundancy masking trade-off}
\label{subsec:masking}

To develop effective self-braking behavior, models learn both when to continue reasoning and when to stop. We investigated different configurations of preserved (unmasked) and masked content to understand this balance. Shown in Table~\ref{tab:masking}, preserving two complete solutions while masking only a few additional sentences yields optimal performance, reducing tokens by 49\% while maintaining 97.4\% of baseline accuracy.

This finding provides two key insights. First, solution repetition serves as a natural termination signal: when a model derives the same answer twice, it learns this is a strong indication to conclude reasoning. Second, we observe an inverse relationship between preserved and masked content: with more preserved reasoning (two solutions), less masked content is optimal; with less preserved reasoning (one solution), more masked content performs better.

This relationship suggests that models require a certain ``reasoning quota'' to develop robust problem-solving capabilities, which can be satisfied either through more preserved reasoning or more exposure to masked reasoning patterns. However, the superior performance of the ``two solutions with minimal masked content'' configuration indicates that clearly delineated, complete reasoning paths provide stronger learning signals than exposure to additional masked content.

\subsection{With vs. without masked redundant thinking}

\label{subsec:mrt}

\begin{wraptable}{r}{0.35\textwidth}

\vspace{-1em}

\caption{MRT comparison. Detailed results shown in Appendix Table~\ref{tab:new1}.}

\label{tab:mrt}

\centering

\small

\begin{tabular}{lcc}

    \toprule

    \textbf{Config.} & \textbf{Acc} & \textbf{\#Tok} \\

    \midrule

    Baseline & 59.36 & 3277 \\

    \midrule

    w/ MRT & \textbf{57.83} & \textbf{1673} \\

    w/o MRT & 58.02 & 2306 \\

    \bottomrule

\end{tabular}
\vspace{-1em}

\end{wraptable}

To assess masked redundant thinking (MRT), we compare SBT-E with and without masked segments. In the ablation variant, trajectories truncate directly without retaining masked content.

Results show removing MRT yields minimal accuracy gain (+0.19\%) but substantial efficiency loss (+37.8\% tokens). Without exposure to masked overthinking patterns, models fail to internalize what constitutes redundancy. Direct truncation provides only implicit stopping signals—models see where reasoning ends but not why. Masked segments serve as explicit negative examples: models observe overthinking without gradient reinforcement, enabling discriminative learning of termination boundaries. This exposure-without-reinforcement principle proves essential for self-regulation while preserving reasoning quality.

\subsection{Impact of $\beta$ on RER-OMR balance}

\label{subsec:beta}

\begin{wraptable}{r}{0.42\textwidth}

\vspace{-12pt}

\caption{Performance with different $\beta$ values for overthink score weighting. Detailed results shown in Appendix Table~\ref{tab:new2}.}

\label{tab:beta}

\centering

\small

\begin{tabular}{lccc}

    \toprule

    \textbf{Method} & $\boldsymbol{\beta}$ & \textbf{Acc} & \textbf{\#Tok} \\

    \midrule

    Baseline & -- & 59.36 & 3277 \\

    \midrule

    \multirow{4}{*}{SBT-E}

    & 0.05 & 56.48 & 1762 \\

    & \textbf{0.1} & \textbf{57.83} & \textbf{1673} \\

    & 0.15 & 56.52 & 1874 \\

    & 0.2 & 55.86 & 1809 \\

    \midrule

    \multirow{4}{*}{SBT-D}

    & 0.05 & 56.24 & \textbf{1678} \\

    & \textbf{0.1} & \textbf{56.66} & 1682 \\

    & 0.15 & 56.21 & 1784 \\

    & 0.2 & 55.74 & 1814 \\

    \bottomrule

\end{tabular}

\vspace{-8pt}

\end{wraptable}

The overthink score (Equation 3) combines structural efficiency $(1-\eta_s)$ and linguistic markers $\kappa_t$ through weighting parameter $\beta$. Table~\ref{tab:beta} examines sensitivity across $\beta \in \{0.05, 0.1, 0.15, 0.2\}$.

Results show $\beta = 0.1$ consistently achieves optimal performance for both SBT variants, with 57.83\% accuracy (SBT-E) and 56.66\% (SBT-D) while maintaining strong efficiency (1673 and 1682 tokens). Lower values ($\beta = 0.05$) over-emphasize linguistic markers, degrading accuracy by 0.4--1.4 points despite minimal token savings. Higher values ($\beta \geq 0.15$) reduce linguistic contribution, causing accuracy drops of 1.3--2.0 points with increased token consumption. This validates prioritizing structural efficiency (90\%) while retaining linguistic signals (10\%) for robust detection without prompt sensitivity.

\subsection{Step-level vs. token-level overthinking detection}
\label{subsec:granularity}

The granularity of overthinking detection, whether operating at the reasoning step level or token level, impacts both the coherence of preserved reasoning and overall model performance. We compared our step-level approach with a token-level alternative using a token efficiency ratio defined as $\eta_t = \frac{FT}{TT}$, where $FT$ represents tokens until first correct answer and $TT$ represents total tokens.

Results in Table~\ref{tab:granularity} demonstrate that step-level detection outperforms token-level approaches, achieving both higher accuracy and lower token usage. This confirms our hypothesis that reasoning coherence is better preserved when entire logical steps are maintained intact. Token-level truncation, while more granular, risks breaking logical units of reasoning, potentially creating disjointed or incomplete thinking patterns that are harder for models to learn from or reproduce effectively.

This finding highlights the importance of respecting the inherent structure of reasoning when developing overthinking mitigation strategies: models benefit from complete logical units rather than more aggressive but potentially incoherent truncation approaches.

\begin{table}[!th]
\centering
    \small
 \begin{minipage}[t]{0.48\textwidth}
    \centering
    \caption{Overthinking detection granularity comparison. Detailed results in Appendix Table~\ref{tab:rawof5.3}.}
    \label{tab:granularity}
    \vspace{1em}
   \begin{tabular}{lrr}
        \toprule
        \textbf{Level} & \textbf{Acc} & \textbf{\#Tok} \\
        \midrule
        Baseline    & 59.36 & 3277 \\
        Step-Level  & \textbf{56.66} & \textbf{1682} \\
        Token-Level & 56.24 & 1753 \\
        \bottomrule
    \end{tabular}
  \end{minipage}
  \hspace{3mm}
  \begin{minipage}[t]{0.48\textwidth}
   \centering
   \small
   \caption{Guiding mode comparison. Detailed results shown in Appendix Table~\ref{tab:rawof5.4}.}
   \label{tab:guidance}
   \vspace{1em}
     \begin{tabular}{lcc}
        \toprule
        \textbf{Guiding Mode} & \textbf{Acc} & \textbf{\#Tok} \\
        \midrule
        Baseline         & 59.36 & 3277 \\
        Natural Language & \textbf{56.66} & \textbf{1682} \\
        Special Token    & 56.61 & 1797 \\
        No Guidance & 56.39 & 1801 \\
        \bottomrule
    \end{tabular}
\end{minipage}
\end{table}

\subsection{Natural language guidance vs. alternative approaches}
\label{subsec:guidance}

A fundamental aspect of Self-Braking Tuning is the mechanism used to signal reasoning termination. We compared our natural language guidance approach (epiphany sentences like ``I've verified my answer, no need to continue...'') with both a special token approach using \texttt{<stop\_overthinking>} and a configuration without any explicit guidance.

Table~\ref{tab:guidance} shows natural language guidance achieves optimal performance (56.66\% accuracy, 1682 tokens). The necessity of explicit termination signals is confirmed by the no-guidance ablation, which degrades both accuracy ($-0.27$\%) and efficiency (+7.1\% tokens). Compared to special tokens (56.61\%, 1797 tokens), natural language guidance maintains comparable accuracy with 6.4\% fewer tokens by leveraging models' existing semantic understanding of logical transitions rather than requiring learning of artificial control conventions.

\section{Conclusion}
In this paper, we propose a novel endogenous approach, Self-Braking Tuning (SBT), to mitigating overthinking in large language models. SBT aims to stimulate the model’s ability to autonomously identify and stop redundant reasoning. We construct a data framework with adaptive reasoning lengths, where overthinking characteristics are extracted through step-level analysis and keyword annotation. Based on this, we design two data generation strategies, SBT-E and SBT-D, to help the model learn when to stop and how to simplify its reasoning process. During supervised fine-tuning, we introduce redundancy masking and epiphany sentences to preserve the reasoning paradigm while enhancing the model’s sensitivity to redundant thought patterns. Experimental results show that SBT models significantly reduce token consumption by 30\%–60\% on multiple mathematical benchmarks, with minimal impact on accuracy. Our work demonstrates the potential for large reasoning models to self-regulate their reasoning and offers a promising direction for more efficient long-chain reasoning.

\newpage
\section*{Acknowledgement}

This work is supported by the National Natural Science Foundation of China (No. 62376245), National Key Research and Development Project (No. 2024YFB3312900), the Key Research and Development Program of Zhejiang Province, China (No. 2024C03255), the Fundamental Research Funds for the Central Universities (226-2024-00170), MOE Engineering Research Center of Digital Library, CCF-Tencent Rhino-Bird Open Research Fund, and ZJU Kunpeng\&Ascend Center of Excellence.

\small
\bibliographystyle{unsrt} 
\bibliography{neurips_2025}
\medskip

\appendix

\newpage

\section{Limitations} 
\label{limit}
While \textit{Self-Braking Tuning} (SBT) has demonstrated its ability to reduce overthinking by enabling models to autonomously regulate reasoning length, several limitations remain:

\begin{itemize}
    \item \textbf{Domain generalization.} 
    Our study focuses primarily on math reasoning tasks (e.g., GSM8K, MATH), which, while structurally rich and sensitive to overthinking, do not cover the full diversity of reasoning challenges. The applicability of SBT to open-ended, commonsense, logical, or multimodal reasoning remains unverified. These domains may exhibit distinct redundancy patterns that require specialized adaptation strategies.

    \item \textbf{Data scalability and adaptivity.}
    Due to computational constraints, our experiments are conducted on a dataset of approximately 92K examples. The impact of scaling to larger datasets (e.g., millions of samples) remains unexplored. Moreover, current data construction strategies (SBT-E/D) rely on fixed threshold parameters for overthinking detection, which may require manual tuning across different tasks and hinder dynamic adaptation.

    \item \textbf{Overthinking signal definition.}
    We use reasoning efficiency ($\eta_s$) and overthink marker ratio ($\kappa_t$) to estimate redundancy, but these metrics may not capture all subtle or latent forms of reasoning utility. Some steps that appear redundant may actually serve a hidden purpose, such as helping the model organize its reasoning or implicitly represent abstract patterns.

    \item \textbf{Interpretability and controllability.}
    While SBT introduces a soft constraint via loss masking, the internal process by which a model decides to terminate reasoning remains opaque. There is no explicit mechanism to trace or control this decision, limiting transparency and reliability---especially in applications requiring high interpretability, such as education, healthcare, or scientific domains.

    \item \textbf{Potential trade-offs in complex tasks.}
    For tasks requiring deep, multi-step reasoning (e.g., theorem proving), prematurely terminating reasoning may risk omitting critical steps. Although SBT preserves accuracy in most cases, it may underperform in scenarios where full-chain reasoning is essential. Adaptive or progressive braking strategies may be needed to better balance efficiency and completeness.

\end{itemize}

Future work could address these limitations by (i) expanding SBT to open-ended and multimodal tasks, (ii) scaling to larger and more diverse datasets, (iii) developing adaptive thresholding and automatic data construction pipelines, (iv) improving multilingual and domain-robust prompt designs, and (v) integrating interpretable or feedback-driven stopping mechanisms that better align with complex reasoning dynamics.

\section{Autonomous reasoning regulation in self-braking models}

A critical distinction of our Self-Braking Tuning framework is that SBT-E and SBT-D are \textit{data construction strategies}, not inference-time control mechanisms. After training, models exhibit autonomous reasoning regulation without external intervention, dynamically adapting reasoning depth to problem complexity.

\subsection{Emergent self-regulation behavior}

To validate that self-braking emerges naturally during inference, we analyzed reasoning patterns of Qwen2.5-Math-7B-Instruct-SBT-E on AIME. Models spontaneously generate epiphany sentences (e.g., ``Wait, I've verified my answer. No need to continue thinking.'') and terminate without external signals.

\begin{table}[H]
\centering
\small
\caption{Analysis of autonomous early termination in SBT-trained models on AIME. Models naturally terminate reasoning in about 50\% of cases, achieving both higher accuracy and 48--51\% token reduction.}

\begin{tabular}{llccc}
\toprule
\textbf{Dataset} & \textbf{Exit Type} & \textbf{\% of Cases} & \textbf{Acc} & \textbf{\#Tok} \\
\midrule
\multirow{2}{*}{AIME24} & Early Exit & 50.83\% & 41.80\% & 5,692 \\
                        & No Early Exit & 49.17\% & 38.98\% & 11,084 \\
\midrule
\multirow{2}{*}{AIME25} & Early Exit & 49.17\% & 41.53\% & 6,483 \\
                        & No Early Exit & 50.83\% & 32.79\% & 12,201 \\
\bottomrule
\end{tabular}
\vspace{12pt}

\label{tab:autonomous_behavior}
\end{table}
\vspace{-12pt}

Table~\ref{tab:autonomous_behavior} shows that self-braking is not enforced through hard constraints. Cases with early termination achieve higher accuracy while using fewer tokens—indicating models learn to recognize when additional reasoning becomes counterproductive.

\subsection{Problem-adaptive reasoning depth}

SBT-trained models autonomously adjust reasoning length based on task difficulty:

\begin{table}[h]
\centering
\small
\caption{Adaptive reasoning depth of Qwen2.5-Math-7B-Instruct-SBT-E across difficulty levels. The 7.3$\times$ variation between GSM8K and AIME25 demonstrates autonomous complexity adaptation without external difficulty indicators.}
\begin{tabular}{lccc}
\toprule
\textbf{Dataset} & \textbf{Difficulty} & \textbf{Avg Steps} \\ 
\midrule
GSM8K   & Easy        & 27.78  \\
MATH500 & Medium      & 51.32  \\
AMC23   & Hard        & 106.40 \\
AIME25  & Very Hard   & 202.23 \\
\bottomrule
\end{tabular}
\vspace{12pt}

\label{tab:adaptive_depth}
\end{table}

\vspace{-12pt}

The substantial variation (27.78 to 202.23 steps) occurs without external difficulty indicators or task-specific prompting, suggesting internalized problem complexity assessment.

\subsection{Interpretation of self-regulation mechanisms}

The training process exposes models to three key signals: (1) preserved reasoning demonstrating sufficient problem-solving depth, (2) natural language cues marking reasoning completion, and (3) masked redundant segments that models observe but do not learn to reproduce. This combination appears to develop an internal representation of ``reasoning sufficiency'' that activates during generation.

We hypothesize that the natural language braking prompts (e.g., ``Wait, I've gotten the same answer multiple times'') serve as anchor points in the representation space, allowing models to recognize similar reasoning states during inference and trigger corresponding termination behavior. The consistency of early termination rates ($\sim$50\% on AIME) across individual problems suggests this behavior generalizes beyond memorization of training examples.

\section{Out-of-domain generalization}
\label{app:ood}
To evaluate whether Self-Braking Tuning generalizes beyond mathematical reasoning, we test SBT-trained models on two out-of-domain benchmarks covering diverse knowledge domains: MMLU-Redux (general knowledge and reasoning)~\cite{gema2025we} and GPQA-Diamond (graduate-level science questions)~\cite{rein2024gpqa}. All models were trained exclusively on math data (OpenR1-Math), with no domain-specific fine-tuning for these benchmarks.

\begin{table}[h]
\centering
\small
\caption{Out-of-domain evaluation on MMLU-Redux and GPQA-Diamond. Models trained exclusively on mathematical reasoning data demonstrate consistent efficiency gains (26--65\% token reduction) across non-mathematical tasks, with accuracy drops typically limited to 1--3\%.}
\begin{tabular}{llcccccc}
\toprule
\multirow{2}{*}{\textbf{Base Model}} & \multirow{2}{*}{\textbf{Method}} & \multicolumn{2}{c}{\textbf{MMLU-Redux}} & \multicolumn{2}{c}{\textbf{GPQA-Diamond}} & \multicolumn{2}{c}{\textbf{Average}} \\
\cmidrule(lr){3-4} \cmidrule(lr){5-6} \cmidrule(lr){7-8}
& & Acc & \#Tok & Acc & \#Tok & Acc & \#Tok \\
\midrule
\multirow{3}{*}{Qwen2.5-Math-1.5B} 
& Baseline & 45.84 & 2,061 & 24.75 & 5,485 & 35.30 & 3,773 \\
& SBT-E & 43.12 & \textbf{1,403} & 25.06 & 3,194 & 34.09 & 2,299 \\
& SBT-D & \textbf{43.28} & 1,566 & \textbf{26.20} & \textbf{3,002} & \textbf{34.74} & \textbf{2,284} \\
\midrule
\multirow{3}{*}{Qwen2.5-Math-7B} 
& Baseline & 67.04 & 3,229 & 41.15 & 8,892 & 54.10 & 6,061 \\
& SBT-E & 65.84 & \textbf{1,927} & 40.40 & \textbf{6,205} & 53.12 & \textbf{4,066} \\
& SBT-D & \textbf{66.39} & 1,998 & \textbf{41.29} & 6,706 & \textbf{53.84} & 4,352 \\
\midrule
\multirow{3}{*}{Llama-3.2-1B} 
& Baseline & 35.62 & 1,933 & 17.99 & 9,321 & 26.81 & 5,627 \\
& SBT-E & 32.24 & 770 & \textbf{24.24} & 3,516 & 28.24 & 2,143 \\
& SBT-D & \textbf{33.12} & \textbf{725} & 23.48 & \textbf{3,157} & \textbf{28.30} & \textbf{1,941} \\
\midrule
\multirow{3}{*}{Llama-3.1-8B} 
& Baseline & 80.53 & 2,481 & 37.31 & 8,918 & 58.92 & 5,699 \\
& SBT-E & 77.46 & \textbf{1,646} & 36.30 & \textbf{5,346} & 56.88 & \textbf{3,496} \\
& SBT-D & \textbf{77.74} & 1,668 & \textbf{36.91} & 6,717 & \textbf{57.33} & 4,192 \\
\bottomrule
\end{tabular}
\vspace{12pt}
\label{tab:ood_results}
\end{table}

\textbf{Cross-domain transferability.} SBT achieves 26--65\% token reduction across non-mathematical benchmarks despite training exclusively on math data, with minimal accuracy degradation (typically 1--3\%). This demonstrates that self-regulation learned from mathematical reasoning generalizes to diverse knowledge domains.

\textbf{Model size effects.} Efficiency gains scale inversely with model size: smaller models (Llama-3.2-1B) achieve 65\% token reduction, while larger models (Llama-3.1-8B, Qwen2.5-Math-7B) show 26--40\% reductions. Notably, Llama-3.2-1B with SBT-E improves GPQA-Diamond accuracy by 6.25 percentage points (17.99\% → 24.24\%), indicating that overthinking mitigation can enhance performance on challenging out-of-domain tasks for smaller models.

\textbf{Architecture consistency.} Both domain-specialized (Qwen2.5-Math) and general-purpose (Llama) models benefit comparably from SBT on out-of-domain evaluation, confirming that self-braking represents a transferable meta-cognitive capability rather than task-specific adaptation.

These results establish that reasoning efficiency patterns learned from mathematics transfer broadly to general knowledge and scientific reasoning, supporting SBT's applicability beyond its training domain.

\section{Extended threshold analysis}
\label{app:extended_threshold}

To comprehensively evaluate the robustness of our overthinking detection mechanism, we conducted an extended threshold analysis spanning $\tau_1 \in \{0.05, 0.1, 0.2, 0.3, 0.4, 0.5\}$. This analysis examines how threshold selection affects both the proportion of samples classified as overthinking and the resulting model performance.

\begin{table}[h]
\centering
\small
\caption{Extended threshold analysis for overthinking score across $\tau_1 \in [0.05, 0.5]$. Results Averaged across GSM8K, MATH500, AIME, and AMC23. Overthink \% indicates the proportion of samples classified as containing overthinking at each threshold.}
\begin{tabular}{lccccc}
\toprule
\textbf{Method} & \textbf{Threshold} & \textbf{Overthink \%} & \textbf{Accuracy} & \textbf{\#Tokens} & \textbf{Reduction} \\
\midrule
Baseline & -- & -- & 59.36 & 3,277 & -- \\
\midrule
\multirow{6}{*}{SBT-E} 
& 0.05 & 75.20 & 55.14 & 1,407 & 57.1\% \\
& 0.1  & 65.40 & 55.59 & 1,427 & 56.5\% \\
& \textbf{0.2}  & \textbf{60.30} & \textbf{57.83} & \textbf{1,673} & \textbf{48.9\%} \\
& 0.3  & 50.20 & 56.70 & 1,755 & 46.4\% \\
& 0.4  & 41.00 & 57.38 & 1,834 & 44.0\% \\
& 0.5  & 2.06  & 57.12 & 2,602 & 20.6\% \\
\midrule
\multirow{6}{*}{SBT-D} 
& 0.05 & 74.20 & 54.90 & 1,215 & 62.9\% \\
& 0.1  & 62.30 & 55.90 & 1,251 & 61.8\% \\
& \textbf{0.2}  & \textbf{62.50} & \textbf{56.66} & \textbf{1,682} & \textbf{48.7\%} \\
& 0.3  & 50.90 & 57.47 & 1,917 & 41.5\% \\
& 0.4  & 40.10 & 57.36 & 1,902 & 42.0\% \\
& 0.5  & 0.19  & 57.09 & 2,696 & 17.7\% \\
\bottomrule
\end{tabular}
\vspace{12pt}
\label{tab:extended_threshold}
\end{table}

\subsection{Threshold regime characterization}

The extended analysis reveals three distinct performance regimes:

\textbf{Aggressive pruning ($\tau_1 < 0.2$).} Low thresholds classify 62--75\% of samples as overthinking, achieving maximum token reduction (57--63\%) but incurring significant accuracy degradation (2.5--4.5 percentage points). At $\tau_1 = 0.05$, the framework removes reasoning content too aggressively, truncating even necessary exploration steps in complex problems. This regime prioritizes efficiency over reasoning completeness.

\textbf{Balanced operation ($\tau_1 \in [0.2, 0.4]$).} This range identifies 40--62\% of samples as overthinking and demonstrates stable performance with $<$1.2\% accuracy variation while maintaining 41--49\% token reduction. The relative insensitivity to exact threshold values within this range indicates robust overthinking detection. Peak performance occurs at $\tau_1 = 0.2$ for both SBT variants, achieving optimal accuracy-efficiency trade-offs (57.83\% / 48.9\% reduction for SBT-E; 56.66\% / 48.7\% for SBT-D).

\textbf{Conservative pruning ($\tau_1 = 0.5$).} High thresholds classify only 0.2--2\% of samples as overthinking, causing the framework to degenerate toward baseline behavior. Token reduction drops to 18--21\% with no compensatory accuracy gains, indicating that insufficient overthinking mitigation fails to realize efficiency benefits.

\subsection{Threshold selection rationale}

We select $\tau_1 = 0.2$ as the default threshold based on three considerations:

\textbf{Performance optimality.} This threshold achieves the best accuracy within the balanced regime while maintaining substantial efficiency gains across both SBT variants and all evaluated benchmarks.

\textbf{Empirical stability.} The plateau in the $[0.2, 0.4]$ range demonstrates that performance degrades gradually rather than catastrophically under threshold perturbations, supporting practical deployability.

\textbf{Dataset composition.} A threshold of 0.2 creates approximately 60\% SBT-processed and 40\% original trajectories, enabling models to learn both when to terminate (from processed examples) and when to continue reasoning (from preserved examples). This balanced exposure prevents over-generalization of braking behavior while establishing clear termination patterns.

These findings establish that SBT's effectiveness stems from systematic overthinking identification rather than aggressive truncation, with the optimal operating point balancing reasoning preservation against redundancy elimination.

\section{Overthink markers}
\label{markers}
In the study of overthinking behaviors, several prior works have highlighted that certain words associated with reflection, hesitation, or backtracking play a critical role in identifying and guiding redundant reasoning processes~\cite{yeo2025demystifying,lu2025retro, fan2025missing,li2025llms}. Building on these insights, we compile a set of common Overthink Markers, including:

\begin{quote}
\small
\begin{tabularx}{\linewidth}{XXX}
Another & Backtrack & But \\
Check & Going back & Hmm \\
Hmmm & However & Hold on \\
Instead of & Just to be thorough & Just to make sure \\
Let me check & Let me just double-check & Let me try another \\
Let me verify & Maybe & Maybe I can consider \\
Maybe I should consider & Might & Not sure \\
Perhaps & Recheck & Retry \\
Trace back & Wait &
\end{tabularx}
\end{quote}

These terms frequently co-occur with behaviors such as repeated verification, alternative hypothesis formulation, or reasoning path retracing, and are therefore treated as linguistic indicators of redundant cognitive load during reasoning. Based on this, we construct the set $\mathcal{M}$ as part of our overthinking detection metric, used to compute redundancy density at the linguistic level and help identify potential efficiency bottlenecks in deep reasoning.

\section{Algorithms of self-braking tuning}
To facilitate reproducibility and offer a transparent view into our data construction pipeline, we provide the formal procedures for the Self-Braking Tuning Exact (SBT-E) and Self-Braking Tuning Dynamic (SBT-D) methods in this appendix. These two strategies represent complementary approaches for curating training data that teach models to regulate their own reasoning length and avoid excessive computation.

SBT-E adopts a uniform truncation scheme, where each reasoning trajectory is truncated at a consistent structural boundary: the Foundation Solution and the first Evolution Solution are preserved, followed by a small masked portion of subsequent reasoning. This approach provides a clean and interpretable signal for identifying the onset of overthinking.

In contrast, SBT-D offers a fine-grained, adaptive strategy that dynamically determines the optimal stopping point for each problem based on the model’s own overthinking scores. It incrementally evaluates reasoning steps, retaining those below a primary overthinking threshold, and masks additional steps that exceed this threshold but remain below a secondary cutoff—effectively preserving problem-specific nuances in reasoning depth.

Crucially, both methods employ loss masking on the redundant segments: the model sees these overthinking patterns during training but receives no gradient updates from them. This enables the model to implicitly recognize and learn to avoid overthinking, without being rewarded for producing verbose or unnecessary steps.

The full algorithmic details are outlined in Algorithms \ref{alg:sbt-e} and \ref{alg:sbt-d}, which serve as a blueprint for implementing the Self-Braking Tuning framework.

\begin{algorithm}
\caption{Self-Braking Tuning Exact (SBT-E)}
\label{alg:sbt-e}
\begin{algorithmic}[1]
\REQUIRE Reasoning trajectory $T$ with Foundation Solution $FS$ and Evolution Solutions $ES = [ES_1, ES_2, ...]$
\ENSURE Modified trajectory $T'$ with preserved and masked segments
\STATE $\text{PreservedSegment} \leftarrow FS + ES_1$
\IF{$|ES| > 1$}
\STATE $\text{MaskedSegment} \leftarrow \text{First 10-20\% of } ES_2$
\ELSE
\STATE $\text{MaskedSegment} \leftarrow \emptyset$
\ENDIF
\STATE $T' \leftarrow \text{PreservedSegment} + \text{MaskedSegment}$ \COMMENT{With loss masking on MaskedSegment}
\RETURN $T'$
\end{algorithmic}
\end{algorithm}

\begin{algorithm}
\caption{Self-Braking Tuning Dynamic (SBT-D)}
\label{alg:sbt-d}
\begin{algorithmic}[1]
\REQUIRE Reasoning trajectory $T$ with steps $[S_1, S_2, ..., S_n]$, thresholds $\tau_1$ and $\tau_2$
\ENSURE Modified trajectory $T'$ with preserved and masked segments
\STATE $\text{PreservedThinking} \leftarrow \text{Foundation Solution from } T$
\STATE $i \leftarrow \text{index of first step after Foundation Solution}$
\WHILE{$i \leq n$ \AND $\text{CalculateOverthinkScore}(\text{PreservedThinking} + S_i) < \tau_1$}
\STATE $\text{PreservedThinking} \leftarrow \text{PreservedThinking} + S_i$
\STATE $i \leftarrow i + 1$
\ENDWHILE
\STATE $\text{MaskedThinking} \leftarrow \emptyset$
\WHILE{$i \leq n$ \AND $\text{CalculateOverthinkScore}(\text{PreservedThinking} + \text{MaskedThinking} + S_i) < \tau_2$}
\STATE $\text{MaskedThinking} \leftarrow \text{MaskedThinking} + S_i$
\STATE $i \leftarrow i + 1$
\ENDWHILE
\STATE $T' \leftarrow \text{PreservedThinking} + \text{MaskedThinking}$ \COMMENT{With loss masking on MaskedThinking}
\RETURN $T'$
\end{algorithmic}
\end{algorithm}

\newpage
\section{Detailed experimental results}
In Section \ref{sec:analysis}, to improve the readability of the main text, we only present the Average results across the datasets. Here, we provide the specific data and evaluation results.
\vspace{-10pt}

\begin{table}[H]
\centering
\caption{
Performance with varying overthink score thresholds. 
Lower thresholds (0.2) achieve optimal efficiency-accuracy trade-offs—up to 49\% token reduction while preserving 97.4\% baseline accuracy—validating aggressive pruning effectiveness. 
Accuracy remains stable across thresholds on simple tasks (GSM8K), while complex tasks (AIME) show marginal accuracy gains at higher thresholds (0.3--0.4) with increased token costs, indicating task-dependent sensitivity to pruning aggressiveness.
}
\vspace{5pt}
\begin{adjustbox}{max width=\textwidth}
\begin{tabular}{L{1.8cm}C{2cm} 
C{0.6cm}C{0.6cm}
C{0.6cm}C{0.6cm}
C{0.6cm}C{0.6cm}
C{0.6cm}C{0.6cm}
C{0.6cm}C{0.6cm}}
\toprule
\multirow{2}{*}{\centering \textbf{Method}} & \multirow{2}{*}{\centering \textbf{Threshold}} 
& \multicolumn{2}{c}{\textbf{GSM8K}}
& \multicolumn{2}{c}{\textbf{MATH500}} 
& \multicolumn{2}{c}{\textbf{AIME}} 
& \multicolumn{2}{c}{\textbf{AMC23}} 
& \multicolumn{2}{c}{\textbf{Average}} \\
\cmidrule(lr){3-4} \cmidrule(lr){5-6} \cmidrule(lr){7-8} \cmidrule(lr){9-10} \cmidrule(lr){11-12}
& & Acc & \#Tok & Acc & \#Tok & Acc & \#Tok & Acc & \#Tok & Acc & \#Tok \\
\midrule
Baseline & - & 85.00 & 514 & 80.25 & 1712 & 16.25 & 7381 & 55.94 & 3503 & 59.36 & 3277 \\
\midrule
\multirow{3}{*}{SBT-Exact} 
& 0.2 & 84.85 & 426 & 77.10 & 1121 & 13.75 & 3101 & 55.63 & 2044 & \textbf{57.83} & \textbf{1673} \\
& 0.3 & 85.16 & 424 & 77.25 & 1113 & 15.63 & 3353 & 48.75 & 2132 & 56.70 & 1755 \\
& 0.4 & 84.73 & 421 & 77.40 & 1130 & 12.71 & 3795 & 54.69 & 1988 & 57.38 & 1834 \\
\midrule
\multirow{3}{*}{SBT-Dynamic} 
& 0.2 & 84.87 & 414 & 77.30 & 1046 & 14.17 & 3381 & 50.31 & 1888 & 56.66 & \textbf{1682} \\
& 0.3 & 84.58 & 410 & 78.00 & 1125 & 12.92 & 3710 & 54.37 & 2422 & \textbf{57.47} & 1917 \\
& 0.4 & 85.07 & 407 & 78.73 & 1187 & 14.38 & 3593 & 51.25 & 2421 & 57.36 & 1902 \\
\bottomrule
\end{tabular}
\end{adjustbox}
\vspace{10pt}
\label{tab:rawof5.1}
\end{table}

\begin{table}[H]
\centering
\caption{
Performance corresponding to different combinations of preserved and masked reasoning. 
The best configuration—preserving two complete solutions while masking only a few redundant sentences—achieves the highest Average accuracy (57.83\%) with 49\% fewer tokens than baseline. 
On simpler datasets like GSM8K, even minimal preservation suffices for learning effective termination, while harder tasks such as AIME benefit more from exposing multiple complete solutions, highlighting that task complexity influences the optimal balance between reasoning exposure and truncation cues.
}
\vspace{10pt}
\begin{adjustbox}{max width=\textwidth}
\begin{tabular}{L{4.5cm}
C{0.6cm}C{0.6cm}
C{0.6cm}C{0.6cm}
C{0.6cm}C{0.6cm}
C{0.6cm}C{0.6cm}
C{0.6cm}C{0.6cm}}
\toprule
\multirow{2}{*}{\centering \shortstack[l]{\textbf{Reservations \&} \\ \textbf{Masked Content}}}
& \multicolumn{2}{c}{\textbf{GSM8K}} 
& \multicolumn{2}{c}{\textbf{MATH500}} 
& \multicolumn{2}{c}{\textbf{AIME}} 
& \multicolumn{2}{c}{\textbf{AMC23}} 
& \multicolumn{2}{c}{\textbf{Average}} \\
\cmidrule(lr){2-3} \cmidrule(lr){4-5} \cmidrule(lr){6-7} \cmidrule(lr){8-9} \cmidrule(lr){10-11}
& Acc & \#Tok
& Acc & \#Tok
& Acc & \#Tok 
& Acc & \#Tok 
& Acc & \#Tok \\
\midrule
Baseline & 85.00 & 514 & 80.25 & 1712 & 16.25 & 7381 & 55.94 & 3503 & 59.36 & 3277 \\
\midrule
1 solution \& A few sentences & 85.23 & 416 & 78.00 & 1103 & 12.71 & 3132 & 51.88 & 2148 & 56.95 & 1700 \\
1 solution \& 1 solution & 85.06 & 432 & 78.60 & 1101 & 13.96 & 3178 & 53.12 & 2076 & 57.69 & 1697 \\
2 solutions \& A few sentences & 84.85 & 426 & 77.10 & 1121 & 13.75 & 3101 & 55.63 & 2044 & \textbf{57.83} & \textbf{1673} \\
2 solutions \& 1 solution & 84.77 & 411 & 77.82 & 1034 & 12.50 & 3092 & 54.69 & 2197 & 57.45 & 1684 \\
\bottomrule
\end{tabular}
\end{adjustbox}

\label{label:rawof5.2}
\end{table}

\begin{table}[H]
\centering
\caption{
Performance comparison of Masked Redundant Thinking mechanism across benchmarks of varying difficulty.
Removing MRT yields marginal Average accuracy gain (+0.19\%) but substantial token increase (+37.8\%), with dataset-dependent effects: simple tasks (GSM8K: +9.6\% tokens) show minimal impact, while complex reasoning tasks experience severe efficiency degradation (AIME: +46.5\% tokens, AMC: +36.4\% tokens).
These results confirm MRT enables efficient termination learning through exposure-without-reinforcement, with benefits scaling with task complexity.
}
\vspace{10pt}
\begin{adjustbox}{max width=\textwidth}
\begin{tabular}{L{3cm}
C{0.6cm}C{0.6cm}
C{0.6cm}C{0.6cm}
C{0.6cm}C{0.6cm}
C{0.6cm}C{0.6cm}
C{0.6cm}C{0.6cm}}
\toprule
\multirow{2}{*}{\centering \textbf{Configuration}}
& \multicolumn{2}{c}{\textbf{GSM8K}} 
& \multicolumn{2}{c}{\textbf{MATH500}} 
& \multicolumn{2}{c}{\textbf{AIME}} 
& \multicolumn{2}{c}{\textbf{AMC}} 
& \multicolumn{2}{c}{\textbf{Average}} \\
\cmidrule(lr){2-3} \cmidrule(lr){4-5} \cmidrule(lr){6-7} \cmidrule(lr){8-9} \cmidrule(lr){10-11}
& Acc & \#Tok 
& Acc & \#Tok 
& Acc & \#Tok 
& Acc & \#Tok 
& Acc & \#Tok \\
\midrule
Baseline         & 85.00 & 514  & 80.25 & 1712 & 16.25 & 7381 & 55.94 & 3503 & 59.36 & 3277 \\
\midrule
w/ MRT          & \textbf{84.85} & \textbf{426}  & \textbf{77.10} & \textbf{1121} & \textbf{13.75} & \textbf{3101} & \textbf{55.63} & \textbf{2044} & \textbf{57.83} & \textbf{1673} \\
w/o MRT         & 85.05 & 467  & 78.30 & 1425 & 14.38 & 4544 & 54.37 & 2788 & 58.02 & 2306 \\
\bottomrule
\end{tabular}
\end{adjustbox}
\vspace{10pt}
\label{tab:new1}
\end{table}

\begin{table}[H]
\centering
\caption{
Performance with different $\beta$ values for overthink score weighting across benchmarks.
While $\beta = 0.1$ achieves best Average performance for both variants (57.83\% SBT-E, 56.66\% SBT-D), optimal values vary by dataset: $\beta = 0.05$ maximizes accuracy on simple tasks (GSM8K) and minimizes tokens on MATH500, while $\beta = 0.1$ dominates on complex tasks (AIME, AMC23).
Lower $\beta$ over-emphasizes linguistic markers with limited efficiency gains, while higher $\beta$ degrades both accuracy and efficiency.
This validates $\beta = 0.1$ as the robust default, balancing structural efficiency (90\%) and linguistic signals (10\%) across task complexities.
}
\vspace{10pt}
\begin{adjustbox}{max width=\textwidth}
\begin{tabular}{L{2cm}C{0.8cm}
C{0.6cm}C{0.6cm}
C{0.6cm}C{0.6cm}
C{0.6cm}C{0.6cm}
C{0.6cm}C{0.6cm}
C{0.6cm}C{0.6cm}}
\toprule
\multirow{2}{*}{\centering \textbf{Method}} & \multirow{2}{*}{\centering $\boldsymbol{\beta}$}
& \multicolumn{2}{c}{\textbf{GSM8K}} 
& \multicolumn{2}{c}{\textbf{MATH500}} 
& \multicolumn{2}{c}{\textbf{AIME}} 
& \multicolumn{2}{c}{\textbf{AMC23}} 
& \multicolumn{2}{c}{\textbf{Average}} \\
\cmidrule(lr){3-4} \cmidrule(lr){5-6} \cmidrule(lr){7-8} \cmidrule(lr){9-10} \cmidrule(lr){11-12}
& & Acc & \#Tok 
& Acc & \#Tok 
& Acc & \#Tok 
& Acc & \#Tok 
& Acc & \#Tok \\
\midrule
Baseline & -- & 85.00 & 514 & 80.25 & 1712 & 16.25 & 7381 & 55.94 & 3503 & 59.36 & 3277 \\
\midrule
\multirow{4}{*}{SBT-E}
& 0.05 & \textbf{84.92} & 433 & 76.45 & \textbf{1089} & 13.13 & 3198 & 51.42 & 2328 & 56.48 & 1762 \\
& 0.1 & 84.85 & \textbf{426} & \textbf{77.10} & 1121 & \textbf{13.75} & \textbf{3101} & \textbf{55.63} & \textbf{2044} & \textbf{57.83} & \textbf{1673} \\
& 0.15 & 84.78 & 448 & 76.80 & 1243 & 13.54 & 3487 & 50.96 & 2318 & 56.52 & 1874 \\
& 0.2 & 84.61 & 441 & 76.05 & 1198 & 12.92 & 3421 & 49.86 & 2176 & 55.86 & 1809 \\
\midrule
\multirow{4}{*}{SBT-D}
& 0.05 & \textbf{84.89} & 421 & 76.50 & 1087 & 13.33 & \textbf{3289} & 50.24 & 1915 & 56.24 & \textbf{1678} \\
& 0.1 & 84.87 & \textbf{414} & \textbf{77.30} & \textbf{1046} & \textbf{14.17} & 3381 & \textbf{50.31} & \textbf{1888} & \textbf{56.66} & 1682 \\
& 0.15 & 84.83 & 437 & 76.95 & 1178 & 13.75 & 3598 & 49.31 & 1923 & 56.21 & 1784 \\
& 0.2 & 84.71 & 429 & 76.30 & 1154 & 13.13 & 3701 & 48.82 & 1972 & 55.74 & 1814 \\
\bottomrule
\end{tabular}
\end{adjustbox}
\vspace{10pt}
\label{tab:new2}
\end{table}

\vspace{-10pt}

\begin{table}[H]
\centering
\caption{
Performance comparison between step-level and token-level overthinking detection. 
Step-level supervision consistently achieves lower token usage across all datasets (e.g., 414 vs. 431 on GSM8K, 1888 vs. 2091 on AMC23), indicating more efficient reasoning truncation. 
Accuracy-wise, the two methods are comparable on GSM8K and MATH500, but step-level clearly outperforms token-level on more challenging tasks such as AIME (14.17\% vs. 11.34\%). 
These results highlight that preserving complete reasoning steps enables better efficiency–accuracy trade-offs, especially for complex problem-solving.
}
\vspace{10pt}
\begin{adjustbox}{max width=\textwidth}
\begin{tabular}{L{3cm}
C{0.6cm}C{0.6cm}
C{0.6cm}C{0.6cm}
C{0.6cm}C{0.6cm}
C{0.6cm}C{0.6cm}
C{0.6cm}C{0.6cm}}
\toprule
\multirow{2}{*}{\centering \textbf{Level}} 
& \multicolumn{2}{c}{\textbf{GSM8K}} 
& \multicolumn{2}{c}{\textbf{MATH500}} 
& \multicolumn{2}{c}{\textbf{AIME}} 
& \multicolumn{2}{c}{\textbf{AMC23}} 
& \multicolumn{2}{c}{\textbf{Average}} \\
\cmidrule(lr){2-3} \cmidrule(lr){4-5} \cmidrule(lr){6-7} \cmidrule(lr){8-9} \cmidrule(lr){10-11}
& Acc & \#Tok 
& Acc & \#Tok 
& Acc & \#Tok 
& Acc & \#Tok 
& Acc & \#Tok \\
\midrule
Baseline     & 85.00 & 514  & 80.25 & 1712 & 16.25 & 7381 & 55.94 & 3503 & 59.36 & 3277 \\
\midrule
Step-Level   & 84.87 & 414  & 77.30 & 1046 & 14.17 & 3381 & 50.31 & 1888 & \textbf{56.66} & \textbf{1682} \\
Token-Level  & 85.09 & 431  & 78.23 & 1088 & 11.34 & 3399 & 50.31 & 2091 & 56.24 & 1753 \\
\bottomrule
\end{tabular}
\end{adjustbox}
\vspace{10pt}
\label{tab:rawof5.3}
\end{table}

\begin{table}[H]
\centering
\caption{
Performance comparison of different guiding mechanisms for reasoning termination. 
Natural language guidance consistently achieves the best efficiency-accuracy trade-off across benchmarks, with 3381 vs. 3647 tokens on AIME and 1888 vs. 2007 on AMC compared to special tokens.
The ablation without any guidance mechanism (No Guidance) demonstrates the necessity of explicit termination signals, showing degraded performance with both reduced accuracy ($-0.27$\%) and increased token consumption ($+7.1$\%) compared to natural language guidance.
These results confirm that semantically aligned, self-reflective cues enable more effective reasoning regulation than explicit control tokens or implicit learning alone.
}
\vspace{10pt}
\begin{adjustbox}{max width=\textwidth}
\begin{tabular}{L{3cm}
C{0.6cm}C{0.6cm}
C{0.6cm}C{0.6cm}
C{0.6cm}C{0.6cm}
C{0.6cm}C{0.6cm}
C{0.6cm}C{0.6cm}}
\toprule
\multirow{2}{*}{\centering \textbf{Guiding Mode}}
& \multicolumn{2}{c}{\textbf{GSM8K}} 
& \multicolumn{2}{c}{\textbf{MATH500}} 
& \multicolumn{2}{c}{\textbf{AIME}} 
& \multicolumn{2}{c}{\textbf{AMC}} 
& \multicolumn{2}{c}{\textbf{Average}} \\
\cmidrule(lr){2-3} \cmidrule(lr){4-5} \cmidrule(lr){6-7} \cmidrule(lr){8-9} \cmidrule(lr){10-11}
& Acc & \#Tok 
& Acc & \#Tok 
& Acc & \#Tok 
& Acc & \#Tok 
& Acc & \#Tok \\
\midrule
Baseline         & 85.00 & 514  & 80.25 & 1712 & 16.25 & 7381 & 55.94 & 3503 & 59.36 & 3277 \\
\midrule
Natural Language & \textbf{84.87} & \textbf{414}  & \textbf{77.30} & \textbf{1046} & \textbf{14.17} & \textbf{3381} & 50.31 & \textbf{1888} & \textbf{56.66} & \textbf{1682} \\
Special Token    & 84.94 & 413  & 77.92 & 1120 & 12.34 & 3647 & \textbf{51.25} & 2007 & 56.61 & 1797 \\
No Guidance      & 84.73 & 421  & 77.05 & 1098 & 13.54 & 3523 & 50.24 & 2162 & 56.39 & 1801 \\
\bottomrule
\end{tabular}
\end{adjustbox}
\vspace{10pt}
\label{tab:rawof5.4}
\vspace{10pt}
\end{table}

\end{document}